\documentclass[10pt,twocolumn,letterpaper]{article}

\usepackage[pagenumbers]{cvpr} 

\usepackage{graphicx}
\usepackage{amsmath}
\usepackage{amssymb}
\usepackage{booktabs}

\usepackage{times}
\usepackage{epsfig}
\usepackage{color}

\usepackage{algorithm}
\usepackage{algorithmic}

\usepackage{multirow}
\usepackage{adjustbox}
\usepackage{bbm}

\usepackage[accsupp]{axessibility} 

\usepackage{tablefootnote}

\newcommand{\rg}[1]{\textcolor{black}{#1}} 
\newcommand{\jl}[1]{\textcolor{black}{#1}} 

\usepackage[pagebackref,breaklinks,colorlinks]{hyperref}

\usepackage[capitalize]{cleveref}
\crefname{section}{Sec.}{Secs.}
\Crefname{section}{Section}{Sections}
\Crefname{table}{Table}{Tables}
\crefname{table}{Tab.}{Tabs.}

\def\cvprPaperID{9} 
\def\confYear{2023}

\begin{document}

\title{DIFT: Dynamic Iterative Field Transforms for Memory Efficient Optical Flow}

\author{
Risheek Garrepalli~~~
Jisoo Jeong~~~
Rajeswaran C Ravindran~~~
Jamie Menjay Lin$^{\dagger}$~~~
Fatih Porikli~~~
\smallskip
\\
Qualcomm AI Research$^{*}$
\\
\smallskip
{\tt\small\{rgarrepa, jisojeon, rajeswar, jmlin, fporikli\}@qti.qualcomm.com}
}
\maketitle

\begin{abstract}
Recent advancements in neural network-based optical flow estimation often come with prohibitively high computational and memory requirements, presenting challenges in their model adaptation for mobile and low-power use cases. In this paper, we introduce a lightweight low-latency and memory-efficient model, Dynamic Iterative Field Transforms (DIFT), for optical flow estimation feasible for edge applications such as mobile, XR, micro UAVs, robotics and cameras. DIFT follows an iterative refinement framework leveraging variable resolution of cost volumes for correspondence estimation. We propose a memory efficient solution for cost volume processing to reduce peak memory. Also, we present a novel dynamic coarse-to-fine cost volume processing during various stages of refinement to avoid multiple levels of cost volumes. 
We demonstrate first real-time cost-volume based optical flow DL architecture on Snapdragon 8 Gen 1 HTP efficient mobile AI accelerator with \textbf{32 inf/sec} and 5.89 EPE (endpoint error) on KITTI with manageable accuracy-performance tradeoffs.

\end{abstract}

{\let\thefootnote\relax\footnotetext{
{
\hspace{-6.5mm} $*$ Qualcomm AI Research is an initiative of Qualcomm Technologies, Inc.} 
\\ \hspace{-6.5mm} $\dagger$ This work was done while at Qualcomm AI Research.
\\ This work has been accepted at Mobile AI Workshop CVPR 2023.
}


\section{Introduction} 

\begin{figure}[t]
\centering
\includegraphics[width=\linewidth]{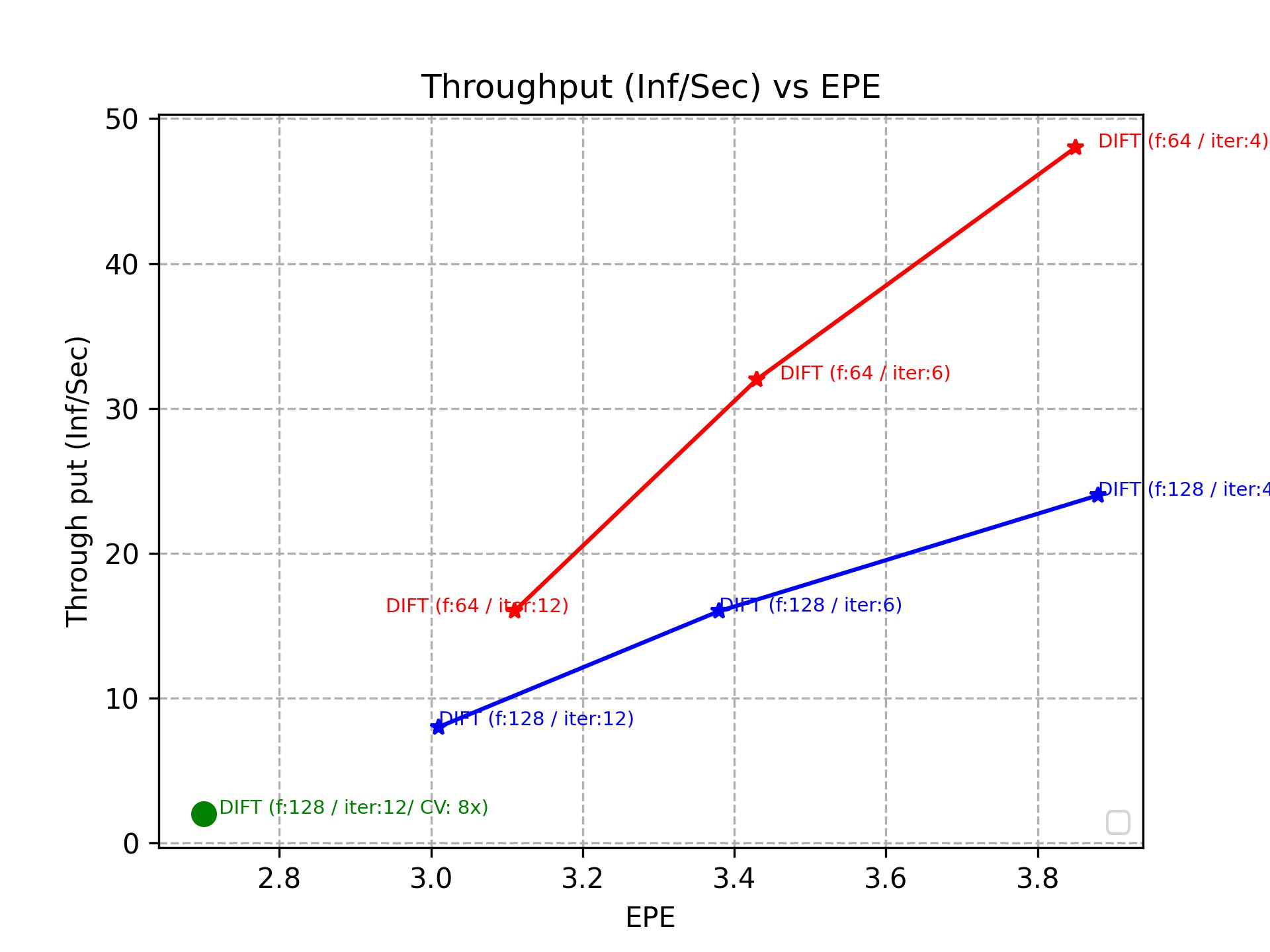}
\vspace{-7mm}
\caption{Accuracy vs. latency comparisons of DIFT over feature dimensions on Snapdragon HTP Platform.
We observe that DIFT with a single-level cost volume at $\frac{1}{16}X$ with feature dimension of 64 ($f=64$) out of 4 iterations achieves an on-target throughput of 48 inf/sec with an EPE of 3.85, while DIFT at $\frac{1}{8}X$ and $f=128$ out of 12 iterations achieves 2 inf/sec with 2.7 EPE.
The RAFT \cite{teed2020raft} baseline is not presented in this figure, as the original RAFT model cannot fit in the memory of commercial smartphones without major modifications in the network model architecture. More details can be found in Table \ref{single_result2}.
}

\label{fig:latency-scatter}
\end{figure}

 



Optical flow is the task of estimating pixel-level correspondences between video frames.
In recent years, starting with \cite{dosovitskiy2015flownet, ilg2017flownet}, deep learning-based optical flow methods have shown remarkable performance while demanding high memory and substantial computation \cite{teed2020raft, xu2022gmflow, jeong2022imposing}. Such computational requirements are not feasible for mobile or other resource-constrained use cases.

Inspired by RAFT \cite{teed2020raft} there are follow up works focusing on efficient approaches \cite{xu2021high, zheng2022dip}, but primary focus of these approaches is to improve performance with some reduction in overall compute.
These approaches too employ cost volumes, requiring significant computational resources and are not directly applicable for mobile or low compute deployment. Design principles of our approach in terms of cost volume processing, number of iterations and resolution of cost-volume can be extended to other modern architectures which employ cost volumes.

\textbf{Global vs Local Cost Volume Processing:} 
If we review various architectures for optical flow, most competitive approaches leverage cost volumes or cross-attention. We can classify approaches by whether they can capture all-pairs similarity measure. 
If approaches have global cost volumes, then it import to understand whether down-stream cost volume processing is based on global operation such as attention, 3D convolution or local operators such as grid-sampling, 2D convolution.

RAFT \cite{teed2020raft} performs grid sampling only in the \textit{local neighborhood} of current estimate of optical flow, but processes information in all-pairs cost-volume i.e, using spatially varying local look-up operation across iterations. Whereas approaches such as GMFlow with the \textit{global} attention or Mobile Stereo \cite{9684954} which applies 3D convolution for stereo correspondence need to operate atomically on global cost volumes, thus requiring much higher computation and peak memory space.

Therefore, \textit{we identify an unique opportunity for magnitude-order complexity reduction in the type of \textit{local} processing particularly with RAFT}. RAFT has significant inductive biases inspired by traditional energy minimization approaches and inspired follow-up approaches that leverage transformers \cite{xu2022gmflow}. We perform additional modifications to RAFT and analyse various design choices and compare to our efficient version DIFT. This helps us in understanding latency vs performance trade-off as illustrated in Figure \ref{fig:latency-scatter} and detailed evaluation is discussed in \ref{ablation_study}. 


Objective of this work is to adopt cost-volume based complex DL architectures and understand critical design choices for mobile use-cases. We introduce \textit{Dynamic Iterative Field Transforms (DIFT)} for optical flow or stereo estimation, a novel memory and computationally efficient architecture for real-time computation on extreme low compute mobile platform which can run with less than 4MB memory while preserving core inductive biases of recent approaches like RAFT \cite{teed2020raft}. 

We identify that single level cost volume \& few iterations give reasonable performance even for large motion (KITTI).
More detailed description of our method can be found in \ref{sec:method2}. Our proposed architecture is evaluated on standard optical flow benchmark datasets \cite{dosovitskiy2015flownet, mayer2016large, butler2012naturalistic, geiger2013vision, menze2015object, kondermann2016hci}.

In summary, inspired by RAFT we propose a novel memory and computationally efficient optical flow architecture (DIFT) for mobile use cases with novel components for better deployment.
\vspace{-0.2mm}
\begin{itemize}
    \setlength\itemsep{-0.3em} 
    \item Introduce a novel coarse-to-fine approach (\ref{coarse2fine}) for correlation lookup with varying cost-volume resolution across iterations.
    \item Extend efficient correlation lookup approach (\ref{jit-method}) from RAFT \cite{teed2020raft} to fit within less than 4MB. 
    \item We introduce an efficient algorithm, \textit{bilinear shift}, (\ref{bilinear_shift2}) which achieves \textbf{$8 \times$} sampling throughput with arithmetic equivalent to the baseline bilinear sampling for HW efficient warping. 
    \item On target deployment and latency evaluation of DIFT, on Snapdragon HTP platform with 8MB Tightly coupled memory (TCM) but also evaluated on platform with less than 4MB.
    
\end{itemize}



\begin{figure*}[t]
\begin{center}


\hspace{-9.4mm}
\includegraphics[width=1\linewidth]{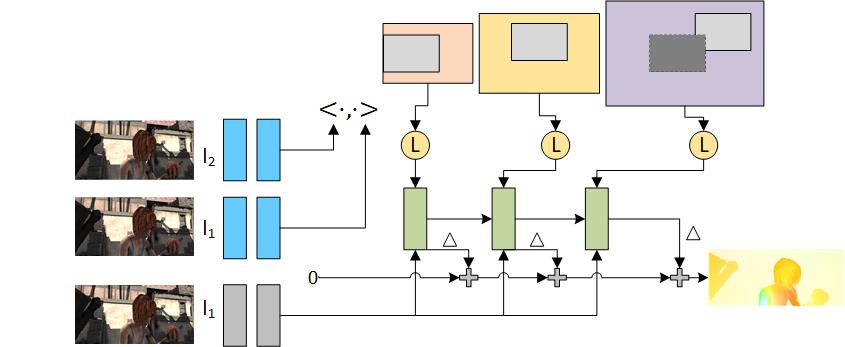}
\end{center}
\vspace{-5mm}
\caption{\textbf{DIFT with Coarse-to-Fine Cost Volumes and Lookup Concatenation}. CNN based encoder applied to images $I_1,I_2$ and additional context Encoder applied to $I_1$. Three different resolutions of cost volumes for various update stages are illustrated in orange, yellow and light purple. Each update block(green) takes output of lookup operator at corresponding cost volume resolution $z_k,r_k$. And overall flow is updated with element-wise sum across iterations as illustrated in bottom of figure. For visual clarity did not include Look-Up Concatenation in the architecture diagram.
}
\vspace{-5mm}
\label{fig_arch}
\end{figure*}

\section{Related Work }
\label{sec:rel}


\textbf{Optical Flow:}
There has been lot of work deep learning based approaches for optical flow estimation, and an active research area. FlowNet \cite{dosovitskiy2015flownet} introduced the first end-to-end CNN based architecture for optical flow estimation, and FlowNet2 \cite{ilg2017flownet} developed a stacked architecture that includes a warping image with optical flow prediction. FlowNet2 shows improves performance by refining optical flow at multiple stages, and such iterative refinement framework is adopted later in \cite{hur2019iterative, teed2020raft}. 

There are various approaches which leverage coarse-to-fine techniques with image or feature pyramids. They compute the large displacement at a coarse level and get refining at the finer level. SpyNet \cite{ranjan2017optical} combined traditional coarse-to-fine image pyramid methods with deep learning, and PWC-Net \cite{sun2018pwc} replaced the fixed image pyramid with learnable feature pyramids. Additionally, PWC-Net conducted the warping operation at the layer level in the network. 
But these approaches process cost volume locally once \& lack capability of iterative refinement architectures. In our work we introduce coarse-to-fine approach for cost volume processing within iterative refinement scheme,
we will describe more in detail in Section \ref{sec:method2}

Recurrent All Pairs Field Transforms (RAFT) \cite{teed2020raft} demonstrated significant performance improvement over previous methods, and has influenced many subsequent research \cite{xu2022gmflow, jeong2022imposing, huang2022flowformer, jeong2023distractflow}. RAFT generates multi-scale and all-pairs field 4D correlation volume from feature extraction outputs and iteratively updates the optical flow estimates through GRU with local grid sampling within 4D correlation volume. 

RAFT has demonstrated good generalization performance and is more robust to adversarial attacks compared to prior works \cite{schrodi2022towards} and has very good inductive biases which is adopted by followup works. Motion is one of fundamental geometric cues applicable in autonomous driving, robotics and many other safety critical applications. Hence a robust and generalizable optical flow estimation is useful not just for down-stream tasks, but also to detect novel objects i.e., openset and Out-of-distribution detection \cite{ruff2021unifying} complementing appearance based features, representational learning and uncertainty measures based approaches \cite{pmlr-v80-liu18e,liu2022pac,garrepalli2022oracle}.
 


\textbf{Memory Efficiency Model:}
Flow1D \cite{xu2021high} decomposed 2D correlation to 1D and DIP \cite{zheng2022dip} proposed patchmatch framework. However, they still require large memory compared to their baseline RAFT algorithm. The Correlation Volume of Flow1D requires $O(HW(H+W))$ memory, which is less than RAFT $O((H+W)^2)$, but depending on resolution, it requires still large memory and pre-computes cost volume. This could also be an interesting direction to consider but we do not consider such decomposition in this work. DIP adopted previous warping method, and it reduced the memory to $O(HW)$. But, DIP computes the correlation volume even at 1/4 downsampled resolution features and has multiple computations to process cost volumes for each update. With low resolution (448 $\times$ 1024), DIP (1.56GB) requires about 3 times more memory compared to RAFT (0.48GB).

\textbf{Modern Optical Flow Architectures:}
Key components of modern optical flow architectures such as RAFT include feature extraction, cost volumes or cross-attention i.e., to use some similarity metric, iterative or multiple stages of flow refinement informed by cost volumes or cross attention, and finally coarse to fine spatial information processing either having multi-scale features and/or multi-scale cost volumes, and having coarse to fine cost volume is potentially useful for capturing large motion.

There are many later works \cite{xu2022gmflow, huang2022flowformer} which adopt core design principles from RAFT by leveraging attention \& transformers while adopting similar design in terms of cost volume or cross attention, (iterative) refinement of flow/correspondence estimation informed by cost volume.

\textbf{Improving Training Pipeline and Data:}
Recent approaches such as Autoflow \cite{sun2021autoflow} and \cite{sun2022makes}, which improve training strategy, or works such as \cite{jeong2022imposing} which leverage augmentations to impose consistency and improve performance, are orthogonal to our work. Our architecture can also leverage such training techniques, but we do not consider such techniques for this work.

\section{DIFT} 
\label{sec:method2}




We introduce DIFT following core design principles of RAFT, an iterative refinement network which consists of a feature extractor, cost volume based iterative refinement scheme with novel components. DIFT adopts a single level of cost volume with potentially varying resolutions. The overall architecture is illustrated in Fig~\ref{fig_arch}, which includes our dynamic coarse to fine lookup, and illustration of sample local grid for a region. 
For DIFT, we do not pre-compute or compute all-pairs cost volume for each update step but can leverage all-pairs information across updates.


Now lets review core components of RAFT and additional modifications for DIFT, 
\textit{we adopt notation from RAFT paper}.
\subsection{Architecture Overview} 
\textbf{Feature Extraction:} We adopt a simple convolutional feature encoder $g_{\theta}$ which is applied on $I_1$ and $I_2$. Output dimension of $g_{\theta}$ is $\mathbb{R}^{H/K \times W/K \times D}$, where `K' is the scale of down-sampling 16 or 8 in our experiments and `D' is the feature dimension, we try $D = 64,128,256$ in our experimental evaluation. Similar to RAFT we also have a separate context encoder applied on base image and output of context encoder is passed to update block. Unlike RAFT for final version we choose $K=16$ to minimize peak memory and overall latency.

\textbf{Cost Volume Computation \& Lookup:} Given images $I_1, I_2$ and image features $g_{\theta}(I_1),g_{\theta}(I_2)$ then \textit{all-pairs} correlation or cost volume $\textbf{C}(g_{\theta}(I_1),g_{\theta}(I_2))$ is computed by applying dot product where each element has its value defined by $C_{ijkl} = \sum_h g_{\theta}(I_1)_{ijh} \cdot g_{\theta}(I_2)_{klh}$.

Let $L_C$ be the correlation lookup, which generates a correlation feature map, is obtained based on current estimate of flow $f^k = (f^1, f^2)$ denote flow in x,y directions, respectively. Then for each pixel $x = (u,v)$ in $I_1$ to obtain correspondence in $I_2$, we define a local neighborhood region around $x$ defined by $\mathcal{N}(x')_r = \{x' + dx | dx \in \mathcal{Z}^2, ||dx||_1 \leq r\}$.

We consider different resolutions of cost volumes $C^1,C^2,C^3$ obtained by pooling on feature maps for correlation lookup,  
or pooling along last two dimension in case of RAFT-small as practiced in \cite{teed2020raft}. For  \textit{coarse to fine lookup}, for each update step we only adopt one level of cost volume $C^i$ and corresponding lookup $L_{C_i}$.

\textbf{Update Block:} Our default update block is adopted from RAFT, lets review components of update block. Each update block takes current flow estimate \textbf{$f^k$}, correlation lookup output, context encoder features.   Update block consists of few pre-processing convolutional layers for processing current flow estimate, correlation features.

Let $z_k, r_k$ be features obtained by after processing current flow $f^k$ and correlation features, respectively. Then based on $z_k, r_k$ and additional input from context encoder and previous hidden state are input to the update block. For \textit{lookup concatenation} we do not just pass $z_k, r_k$ but also $z_{k-1},z_{k-2}$ and $r_{k-1},r_{k-2}$ also as inputs to update block.


\subsection{Coarse-to-Fine Lookup}
\label{coarse2fine}
For DIFT, we adopt a single level cost volume per-iteration and we choose $1/16$ as resolution for cost-volume (CV). To reduce overall latency while aiming to capture large motion, we introduce \textit{Coarse-to-fine strategy for choosing resolution of cost volume} to have a varying effective receptive field across iterations. We achieve this by adopting a coarser resolution for CV for initial iterations, with an intuition that these initial steps should provide good initialization for further refinement. For later iterations, increase the resolution of cost volume so that now the refinement can focus on finer correspondence. And if needed, we can also have one/two iterations of coarser cost volume resolution after finer refinements to capture if there are any regions with larger displacements. 

To dynamically vary cost volume, we perform average pooling on the encoder's feature maps to modify the resolution of cost volume. Figure \ref{fig_arch}, demonstrates varying cost volume resolutions across iterations. We can also consider a method where we vary the lookup radius and potentially consider irregular sampling with the same cost volume resolution across updates. But such irregular sampling or larger neighborhood sampling is inefficient on hardware, and hence we do not investigate this approach.

\textbf{Weight Sharing:}
As we have varying receptive fields w.r.t cost volume, we do not share parameters of update block 
across all iterations, but weight sharing is still present for update blocks which process same resolution of cost volume.


\subsubsection{Look-Up Concatenation}

Usually update block within RAFT only has local neighborhood information within cost volume, but GMFlow \cite{xu2022gmflow} and other works have shown performance improves with global processing of cost volume, but global processing would require significant memory. To retain more information than local neighborhood information captured in lookup operation.
We concatenate the output of lookup operation, previous flow estimate (could also be equivalent to position encoding) over past few iterations as additional input to update block. 

For simplicity we did not include concatenation part in \ref{fig_arch} but darker square block illustrates additional region of correlation volume which can be given as additional input capturing more information than local look-up with less computation \& memory.

Let $L_{C_t}$ be the lookup operator at iteration `k' which returns correlation features and $z_k,r_k$ based current estimate of flow $f^k$. And similarly let  $z_{k-1},r_{k-1}$ are obtained by previous lookup operations $L_{C_{k-1}}$ and  $z_{k-2},r_{k-2}$ from $L_{C_{k-2}}$. When we adopt concatenation within lookup our update block does not just take $z_k,r_k$ but also $z_{k-1},r_{k-1}$ \& $z_{k-2},r_{k-2}$. 


\subsection{Just-in-Time Construction \& Lookup}
\label{jit-method}


Extending a
\jl{computation alternative that samples for each pixel only by the required feature grids based on corresponding neighborhood as discussed in RAFT \cite{teed2020raft}, we further decompose the linear-complexity construction and look-up operations in our \textit{just-in-time} (JiT) approach} 
in order to achieve low peak memory against the tightly-coupled memory (TCM) (4MB) constraints on typical smartphones.

Even during such on-demand lookups we technically need to construct a 3D cost volume based on \jl{a fix-sized} look-up radius (R) and such approach would need $O(N \times 2R)$ memory complexity with $N = (H/16 \times W/16)$, 
\jl{which may not in practice
be feasible to fit in a memory} smaller than 2MB. 

Specifically, to ensure a system of balanced pipelining, which involves memory read/write accesses and neural network processing on our target hardware, we optimize parameters including $N_{slice}$ i.e., the number of tiles in $g_{\theta}(I_1)$ decomposition and $R$ for the radius of look-up range in either direction for each pixel. DIFT with the JiT design successfully achieves a peak memory at only $2MB$ with \{$((H/16 \times W/16) \times (2R))/N_{slice}$\}. In the case of Sintel, we adopt $N_{slice} = 56$ for DIFT at $1/16$ cost volume resolution.


Instead of processing entire feature map $g_{\theta}(I_1)$ at once, \textit{we process it few pixels at a time in a sliding window fashion} and based on neighborhood $\mathcal{N}(x')_r$ 
,retrieve corresponding features from $g_{\theta}(I_2)$ to construct correlation volume for current slice, and then aggregating to generate overall correlation volume for current iteration. In our implementation \textit{we decompose real-valued and sub-pixel part of current flow estimate}, to first use real-valued neighborhood to construct cost volume in a sliding window approach and later adjust for sub-pixel part using bilinear shift approach discussed in next section.



\begin{table*}[t]
\centering
\caption{
Analysis for RAFT \& DIFT on Levels of cost volume, Cost volume resolution, feature encoder dimension,  inference time iterations and corresponding latency. 
($L^{CV}$: Cost Volume Levels, $R^{CV}$: Cost Volume Resolution (Downsampling), $\#F$: Encoder Feature Dimension, $\#I$: Test iterations and $\#radius$: lookup radius). 
C+T represents model trained on Flyingchairs and FlyingThings \& VK indicates fine-tuned on Virtual KITTI for KITTI evaluation only.
In below table best latency-aware version is highlighted with bold, overall best performing DIFT model is denoted via underline.
}
\label{single_result2}
\vspace{-2mm}
\adjustbox{max width=0.98\textwidth}
{


\begin{tabular}{|cccc|cc|cc|c|c|cc|cc|c|c|}
\cline{1-15}
\multicolumn{9}{|c|}{RAFT} & & \multicolumn{5}{|c|}{DIFT (Ours)}\\
\cline{1-15}
\multirow{2}{*}{$L^{CV}$}& \multirow{2}{*}{$R^{CV}$}& \multirow{2}{*}{$\#F$}& \multirow{2}{*}{$\#I$} & \multicolumn{2}{c|}{Sintel (train-EPE) $\downarrow$}  & \multicolumn{2}{c|}{KITTI (Train)} & On-target  
&& \multicolumn{2}{c}{Sintel (train-EPE) $\downarrow$}  & \multicolumn{2}{c|}{KITTI (Train)} & On-target  \\
&&&& Clean & Final & EPE$\downarrow$ & F1-all $\downarrow$ & (ms (Inf/s)) & & Clean & Final & EPE$\downarrow$ & F1-all $\downarrow$ & ms$\downarrow$ (Inf/s)$\uparrow$  \\
\cline{1-15}

\multicolumn{15}{|c|}{Cost Volume Level (C+T)}\\
\cline{1-15}
4 & \multirow{4}{*}{$\frac{1}{16\text{x}}$}&\multirow{4}{*}{ 128}&\multirow{4}{*}{ 12} & 2.65 & 3.94 & 11.87 & 38.02 & \multirow{4}{*}{N/A} &  &N/A &N/A &N/A &N/A &N/A \\ 
3 &&& &  N/A &N/A &N/A &N/A &&& 3.20 & 4.17 & 12.54 & 43.07 & 496.7 (2)\\
2 &&& & 2.67 & 3.81 & 11.23 & 38.14 & & &    3.07  & 4.18 & 12.67 & 41.15 & 248.3 (4)\\
1  &&&& 2.69 & 3.76 & 10.63 & 38.38 & & &    3.01 & 4.32 & 12.21 & 42.06 & 124.9 (8)\\
\cline{1-15}

\multicolumn{15}{|c|}{Cost Volume Resolution (C+T)}\\
\cline{1-15}
\multirow{2}{*}{1} & $\frac{1}{8\text{x}}$  & \multirow{2}{*}{ 128}& \multirow{2}{*}{12} & 2.34 & 3.42 & 9.06  & 28.20 & \multirow{2}{*}{N/A} &  &  \underline{2.70} & 3.93 & 10.33 & 31.47 & \underline{498.7 (2)} \\
& $\frac{1}{16\text{x}}$ && & 2.69 & 3.76 & 10.63 & 38.38 & & &  3.01 & 4.32 & 12.21 & 42.06 & 124.9 (8)\\
\cline{1-15}

\multicolumn{15}{|c|}{Feature Dimension (C+T)}\\
\cline{1-15}
\multirow{2}{*}{1}&\multirow{2}{*}{$\frac{1}{16\text{x}}$} &128 & \multirow{2}{*}{ 12} & 2.69 & 3.76 & 10.63 & 38.38 & \multirow{2}{*}{N/A} & &  3.03 & 4.32 & 12.21 & 42.06 & 123.2 (8) \\
&& 64 & & 2.67 & 3.81 & 10.52  & 37.35 & & &  3.11 & 4.19 & 12.87  & 43.83 & 62.5 (16)\\
\cline{1-15}

\multicolumn{15}{|c|}{Number of Iterations (C+T/VK)}\\
\cline{1-15}
\multirow{3}{*}{1}&\multirow{3}{*}{$\frac{1}{16\text{x}}$} &\multirow{3}{*}{128} & 4 & 3.31 & 4.41  & 6.55 & 24.76 & & &  3.88 & 5.08  & 7.25 & 30.06 & 41.7 (24) \\
&&  & 6 & 2.95 & 4.03 & 5.91  & 22.29 & &  &  3.38  & 4.63  & 6.45  & 26.69 & 62.5 (16)\\
&&  & 12 & 2.68 & 3.76 & 5.11  & 24.33  & &  &  3.01  & 4.32 & 5.67  & 24.33  & 125 (8)\\
\cline{1-15}

\cline{1-15}
\multicolumn{15}{|c|}{Proposed Method (Coarse2Fine) (C+T/VK)}\\
\cline{1-15}
\multirow{3}{*}{1}&\multirow{4}{*}{$\frac{1}{16\text{x}}$} &\multirow{3}{*}{128} & 4 & &&&&&&   3.61   & 4.71 & 6.66 & 30.41 & 42.1 (24) \\
\multicolumn{3}{|c}{}& 6 &&&&&&&   3.38  & 4.44 & 5.83 & 28.05 &  62.9 (16) \\
\multicolumn{3}{|c}{}& 12 &&&&&& &  3.09 & 4.13 & 5.56 & 25.76 &  126.3 (8) \\
1  & & 64 & 6 &&&&&& &  \textbf{3.44} & 4.57 & \textbf{5.89} & 26.81 &  \textbf{32.8 (32)} \\
\cline{1-15}

\multicolumn{15}{|c|}{Proposed Method (Coarse2Fine + Concat) (C+T/VK)}\\
\cline{1-15}
\multirow{3}{*}{1}&\multirow{3}{*}{$\frac{1}{16\text{x}}$} &\multirow{3}{*}{128/} & 4 & &&&&&&   3.64 & 4.91 & \textbf{6.55} & 31.1 & 45.45 (22)\\
\multicolumn{3}{|c}{}& 6 &&&&&& &  3.54 & 4.62 & 6.35 & 30.40 &  71.4 (14) \\
\multicolumn{3}{|c}{}& 12 &&&&&& &  3.04 & 4.15 & 5.51 & 25.76 & 142.85 (6) \\
\cline{1-15}

\end{tabular}
}
\centering
\end{table*}


\subsection{Bilinear Shift}
\label{bilinear_shift2}
Bilinear Sampling is used within each iteration of GRU update to perform warping based on current estimate of flow.
Bilinear sampling involves a series of pointwise grid operations followed by series of bilinear interpolation operations based on current estimate of flow. 
This operation is not efficient on hardware as it cannot leverage vectorized execution and hence adds significant bottleneck to latency. To alleviate this problem, we propose a mathematically equivalent Bilinear shift which replaces grid sampling with vectorizable element-wise operators, and we assume grid input is uniform which can work for dense optical flow estimation.

Based on a shift along x and y dimension i.e., $\Delta x, \Delta y$, the grid sample operation can be decomposed into a bilinear operation associated with a 2D shift, 
and such operations can be mapped to vectorizable elementwise operators on hardware, and can also allow flexibility w.r.t compilation/run-time as it can allow further decomposition into smaller operators/tiling. 


Algorithm \ref{bilinear_shift_algo} describes overall Bilinear shift operation given input tensor $T$, the x-shift ($\Delta x$) and y-shift ($\Delta y$).
For bilinear shift, first input tensor $T$ is split into two equal parts indexed by one pixel offset along x direction and then interpolated using x-shift $\Delta x$ and $(1 - \Delta x)$. Then an analogous operation is repeated along y direction.\\
We compare bilinear shift with baseline grid sampling on our hardware simulation platform to analyze latency. For a given input dimension, bilinear shift has a throughput of 12277.3inf/s (81.45$\mu$s) whereas grid sample has throughput of 1483.2 inf/sec (674.22$\mu$s) i.e., \textit{$8\times$ improvement in throughput (inf/sec)}\\ 

\setlength{\textfloatsep}{10pt} 
\begin{algorithm}[t]
Let input tensor to bilinear shift be $T$ \\
Let $T_{x0}, T_{x1}$ be sliced tensors in width ($X$) axis \\
Let $T_{y0}, T_{y1}$ be sliced tensors in height ($Y$) axis \\
Let $l_{T_x}, l_{T_y}$ be length of Tensors in $X,Y$ axis \\
\quad $T_{x0} = T[0: l_{T_x} - 2]$ and $T_{x1} = T[1: l_{T_x} - 1]$ \\
\quad $T_{x0} = (1- \Delta x) \times T_{x0}$ \\
\quad $T_{x1} = (\Delta x) \times T_{x1}$ \\
\quad $T_2 = T_{x0} + T_{x1}$ \\
\quad $T_{y0} = T_2[0: l_{T_y} - 2]$ and $T_{y1} = T_2[1: l_{T_y} - 1]$ \\
\quad $T_{y0} = (1- \Delta y) \times T_{y0}$ \\
\quad $T_{y1} = (\Delta y) \times T_{y1}$ \\
\quad $T_{out} = T_{y0} + T_{y1}$ \\
\textbf{return} $T_{out}$
\caption{Bilinear Shift}
\label{bilinear_shift_algo}
\end{algorithm}

\vspace{-5mm}

\section{Experimental Setup}
\label{sec:exp}

\subsection{Datasets}
\label{datasets}
We evaluate our approach on common optical flow benchmarks,  Sintel (S) \cite{butler2012naturalistic}, KITTI (K) \cite{geiger2013vision, menze2015object}. For evaluation on Sintel we use checkpoint pre-trained on 
on FlyingChairs (C) \cite{dosovitskiy2015flownet} $+$ Flying Things(T) \cite{mayer2016large}, following a common protocol. And for KITTI evaluation, we start with pre-train on (C+T) and fineune (T) checkpoint additionally on Virtual KITTI2 (VK) \cite{cabon2020virtual}.

\subsection{Architecture Details}

\jl{We build our DIFT model on top of the RAFT \cite{teed2020raft} baseline \footnote{\url{https://github.com/princeton-vl/RAFT}} along with its original hyperparameters}. 
We adopt RAFT-small CNN architecture for Image encoders as starting point, we add one additional convolutional layer to further down-sample input to $1/16x$ image resolution for cost volume processing.  We choose lookup radius to be `3' for all our experiments based to maximize throughput on Snapdragon HTP hardware. And we don't adopt convex up-sampling.


\subsubsection{Weight Sharing of Update Block} 
For baseline single level cost volume variant of DIFT we adopt update block of RAFT-small and have shared weights across iterations i.e., we use a convolutional GRU based on GRU Cell. But for coarse to fine variant, we observe that weight sharing across iterations actually performs worse and \textit{we share weights dependent on resolution of cost volume} across iterations. 

For concatenation of lookup output, we take output of convolutional pre-processing layers before GRU cell within update block from previous iterations as additional information so if required GRU block can leverage such additional context. 
Even though we vary the cost-volume resolution across iterations but as the reference optical flow and cost-volume resolution is fixed and only the relative fidelity of each update operation changes the training is stable and convergence is fast on-par with RAFT compared to previous  DL based approaches such as PWC-Net or FlowNet2.

\subsection{Implementation Details}
We pre-train DIFT for 100k iterations on FlyingChairs (C) \cite{dosovitskiy2015flownet} for 100k steps followed by training on FlyingThings3D (T) \cite{mayer2016large} initialized with previous checkpoint. We adopt batchsize of 12 with AdamW \cite{loshchilov2018decoupled} optimizer for both the datasets. When trained on Virtual KITTI \cite{cabon2020virtual} we adopt RAFT fine-tuning hyper-parameters for KITTI. 
We do not perform any finetuning on Sintel or KITTI and both the datasets are only used for evaluation.
Also, to have a fair comparision with DIFT we also adopt a maximum of 12 iterations/GRU Update steps for RAFT-small.
But we also evaluate both DIFT \& RAFT-small variants after 4,6 and 12 updates for various ablation study as latency is a key constraint for mobile deployment.


\section{Results}

Primary goal of our work is to obtain a good real-time optical flow, stereo or broadly correspondence estimation architectures with good inductive biases, generalization ability for mobile platforms. In this section we want to understand what are key bottlenecks \& critical design choices  in modern cost-volume based DL based architectures such as \cite{teed2020raft}  for mobile or low compute settings. 
We analyse key design choices as discussed in \ref{ablation_study} with in framework inspired by RAFT \cite{teed2020raft} to understand what matters in correspondence estimation for mobile/low-compute settings.

\subsection{Ablation Study} 
\label{ablation_study}



To understand \textit{performance \textit{vs} latency trade-off}. In section \ref{sec:method2}, we discussed our architecture and motivation for our design choices.  Here we present results for representative combinations of design choices and provide guideline recommendation for architecture exploration for correspondence estimation.

\textbf{Coare-to-Fine Lookup:} To understand effectiveness of our proposed approach we can compare Coare-to-fine variants with fixed resolution of cost volume across iterations. From Table \ref{single_result2} we can observe that both for fewer iteration setting coarse-to-fine lookup gives significant performance boost. At 4 iterations, our approach improves performance form $7.25 \to 6.66$ EPE i.e., 0.59 boost in EPE over baseline corresponding DIFT. Similarly \textit{$Delta$ 0.62 boost in EPE} for 6 iterations with final EPE of 5.83 only 0.16 EPE less than 12 iteration version of DIFT. 
In addition to Coarse-to-fine approach lookup-concatenation further improves the performance by 0.11 EPE for 4 iteration setting. 

Overall we observe that our Coarse-to-fine approach consistently improves performance \textit{without additional increase in latency} and is especially effective for large motion setting such as KITTI compared to Sintel, where even a single level cost volume would allow to provide good initialization


\textbf{Number of Levels of Cost Volume:}
We observe that though additional levels of cost-volumes does not necessarily improve performance significantly but adds significant cost to latency. From table \ref{single_result2} we can observe that when go from 1 level cost-volume to 3-levels in case of DIFT, EPE actually gets from 3.01 to 3.20 in case of sintel-clean, similarly it worsens in case of KITTI. But performance improves marginally in case of Sintel Final for DIFT and also in case of RAFT-small variants. But as the performance of single-level cost volume is good-enough it might be worthwhile to investigate such approaches towards efficient solutions.

\textbf{Cost Volume Resolution:} 
In our experiments we find that resolution of cost-volume is one of significant design choices w.r.t performance. When we go from $H/8 \times W/8$ to $H/16 \times W/16$ we can observe that EPE drops from 3.93 to 4.32 i.e., a 0.39 loss of EPE on sintel Final in case of DIFT. But as the latency is also reduced by 4x, we choose to pursue $1/16x$ resolution. 

\textbf{Effect of Iterations:}
Based on Table \ref{single_result2} we can observe that though increasing iterations improve performance, relative improvement is marginal compared to additional compute cost from 5.83 to 5.56 EPE for double the latency in case of KITTI.

\textbf{Feature Dimension:}
In our experiments we find that reducing feature dimension to as low as 64 still gives reasonably good performance.  This helps us get to close to real-time solutions by reducing the feature dimension.

\textbf{Overall Performance:}
Compared a best-performing DIFT with 12 iterations and one $H/8 \times W/8$ cost-volume resolution we get a \textbf{16x} boost in latency with relatively minimal performance drop $2.70$ (2inf/sec) $\to 3.46$ (32 inf/sec) EPE on sintel clean where we adopt a coarse-to-fine variant of DIFT with 6 iterations and feature-dimension/number of encoder channels is 64. In case of KITTI we can get ~5.89 EPE with real-time latency of 32 inf/sec. 

\textit{To the best of our knowledge this is first real-time demonstration of cost-volume based deep learning solution} for extreme low-compute mobile platforms. From \ref{fig:qualitative} we can observe that DIFT output is on-par with RAFT-small while being significantly faster.


\begin{figure*}[t]
\begin{center}$
\centering
\begin{tabular}{c c c c}
\textbf{Ground Truth} & \textbf{RAFT-Small (Original)} & \textbf{RAFT-Small*} & \textbf{DIFT (Ours)}
\\ 
\hspace{-0.2cm} \includegraphics[width=4.2cm, height=2.2cm]{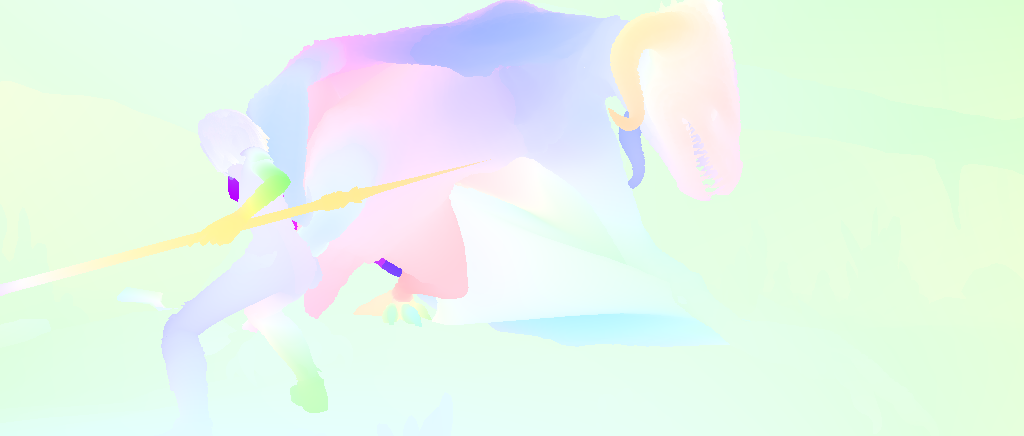} & 
\hspace{-0.4cm} \includegraphics[width=4.2cm, height=2.2cm]{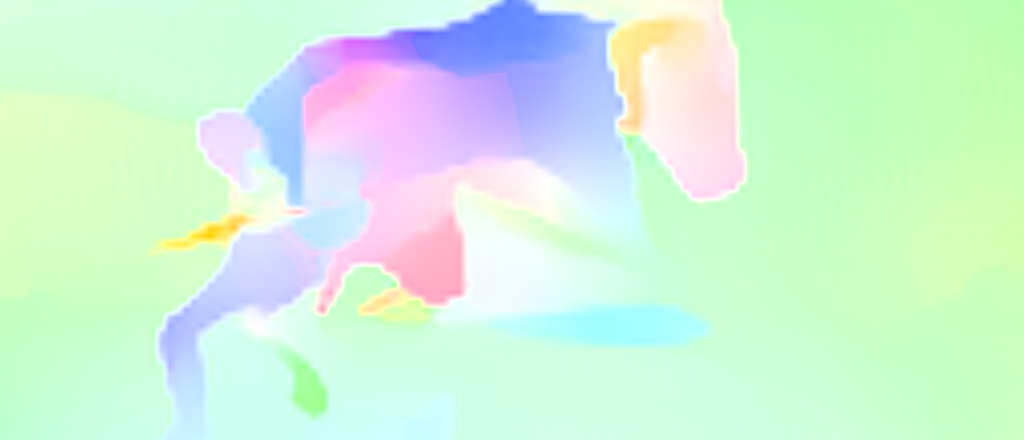} & \hspace{-0.4cm} \includegraphics[width=4.2cm, height=2.2cm]{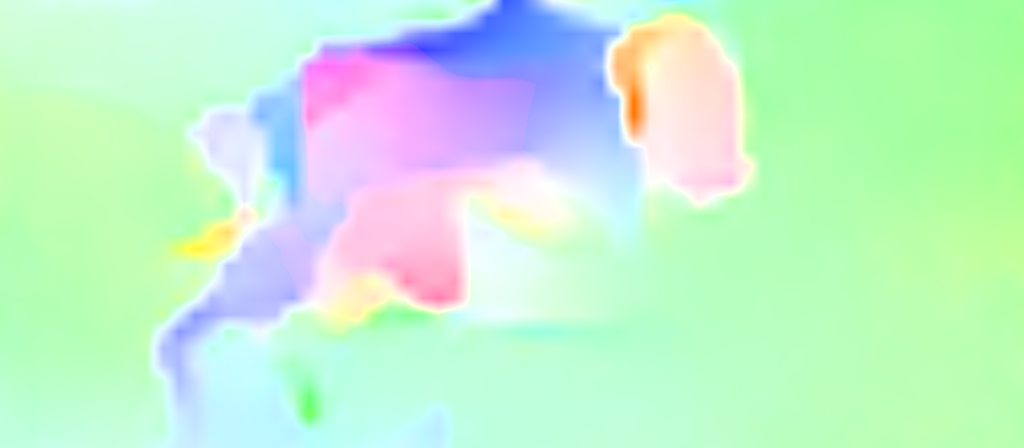} &
\hspace{-0.4cm} \includegraphics[width=4.2cm, height=2.2cm]{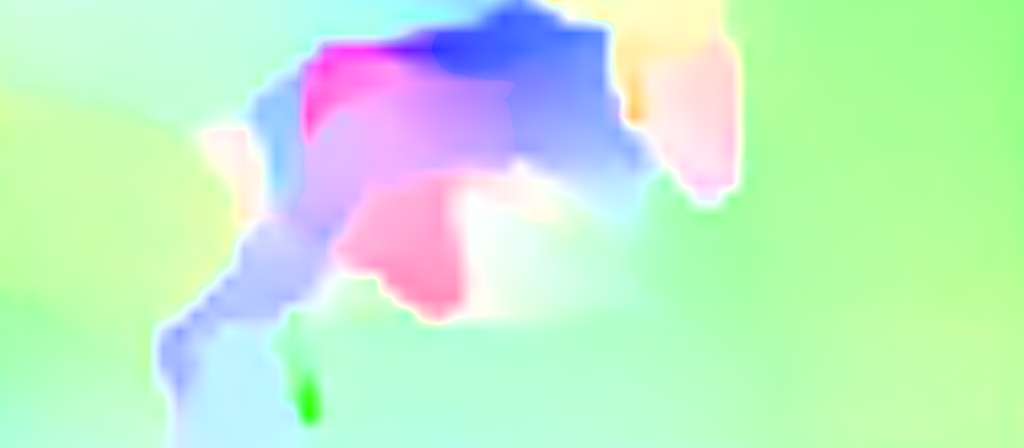}

\end{tabular}$
\end{center}
\caption{ In this figure we compare optical Flow output on a Sintel test-case comparing DIFT to RAFT Variants. RAFT-small is default version with 4 level cost volume, 24 GRU iterations and $1/8 \times$ cost-volume resolution. Where is DIFT and RAFT-small* are single-level cost-volume versions with 6 GRU iterations.}
\vspace{-3mm}
\label{fig:qualitative}
\end{figure*}

\subsection{On-target Evaluation}

In our experiments, we also evaluate DIFT on Snapdragon 8Gen 1 HTP (power efficient mobile AI accelerator for Neural Networks) to understand latency of different design choices. We have developed and optimized a deployment pipeline 
Snapdragon® 8 Gen 1 HTP
accelerator platform, to leverage maximum out of hardware for this model. \rg{We adopt INT8 quantization (W8A8) based on AIMET quantization toolkit \cite{siddegowda2022neural} and QNN-SDK \footnote{\url{https://developer.qualcomm.com/software/qualcomm-ai-stack} }from Qualcomm® AI Stack.} \footnote{ Snapdragon and Qualcomm branded products are products of Qualcomm Technologies, Inc. and/or its subsidiaries.}

Depending on target platform few details might influence different final performance, but our experiments and observations should be informative and translate to other platforms.

\subsection{Analysis}

From Table \ref{single_result2}, we can observe that latency is largely dependent on \textit{size/memory of cost volume}, as the bottleneck is largely w.r.t memory access for such large tensors (cost volumes) after optimizing inference pipeline. Based on these on-target latency evaluations, we can infer following dependencies w.r.t key design choices
\begin{itemize}
 \setlength\itemsep{-0.2em} 
    \item Latency is proportional to levels of Cost Volume as the number of lookups we perform is dependent on number of levels of cost-volume per GRU Update.
    \item Latency is proportional on maximum resolution  of cost volume as this determines the memory transfer and the also GMACs within lookup. 
    \item Latency is proportional to number of iterations , as this also effects total number of lookups per inference.
    \item Latency is proportional to feature/channel dimension as this would also effect memory size of cost volume.
    
    
\end{itemize}

In summary, we observe that \textit{latency is proportional to number of lookup operations and memory requirement of each lookup,} as this is inherently a memory bounded operation and hence latency is significantly determined by peak-memory of hardware module and cost-volume size. 

If there is an efficient way to decompose cost-volume or cross-attention construction and processing more broadly then that would put less constraints on our network architectures.

We can observe from the Table~\ref{single_result2} that, as iterations decrease from $12 \to 4$ throughput triples from 8 inf/sec $\to$ 24 inf/sec, or latency drops from 126.3ms to 42.1ms. Reducing feature dimension from 128 to 64 at 4 iterations further improves throughput from 24 inf/sec to 48inf/sec or 20.8ms latency. 

\subsubsection{Peak Memory}
In this section, we discuss the effect of peak memory on overall latency, as determined by number of lookups which is dependent on size of cost volume and number of levels of cost-volumes for each GRU iteration. 

From table \ref{peak_memory} we can observe that $1/8x$ variant of RAFT needs peak memory of 56.3 MB which is a significant challenge for mobile platforms as they have very limited TCM memory, often less than 10MB. As further decomposition of JiT in DIFT allows to decompose overall operation, if needed DIFT supports higher resolution of cost-volume. 

In case of Sintel, even after 16x down-sampling instead of 8x (for  efficient latency), and \textbf{radius =  3}, we will have 440 x 1024 original image down-sampled, then effective cost volume including padding would result to $28 \times 64 \times (2 \times (radius = 3+1)) ^2 \times 128 = 28 \times 64 \times 64 \times 128 = 14,680,064/1024 = 14.3 MB$ and similarly at 8x and $radius = 3$ its 56.32MB. For DIFT with 4 levels of cost volume (16x), which is similar to RAFT it would take approximately \textbf{($ ~124.9 \times 4 = 499.6ms$)} for one iteration which is equivalent to DIFT at 8x down-sampling but less effective.

\begin{table}[]
\caption{Peak Memory Comparison DIFT \& RAFT}
\label{peak_memory}
\vspace{-2mm}
\adjustbox{max width=0.5\textwidth}
{
\begin{tabular}{llllll}
\cline{1-5}
\textbf{Model} 
& \begin{tabular}[c]{@{}l@{}}\textbf{Down}\\ \textbf{sampling}\end{tabular} 
& \begin{tabular}[c]{@{}l@{}}\textbf{Peak}\\ \textbf{Memory}\end{tabular} 
& \begin{tabular}[c]{@{}l@{}}\textbf{$L^{CV}$} \\  \end{tabular} & \begin{tabular}[c]{@{}l@{}}
\textbf{Latency for} \\ \textbf{12 Iterations}\end{tabular} &  \\
\cline{1-5}
DIFT  & 16x        & $<1MB$                    &   1                       &    124.9 ms                    \\
DIFT  & 16x        & $<1MB$                    &   4                      &    501.3 ms                    \\
RAFT  & 16x        & $14.3 MB$                    & 4                      &  N/A                       \\

DIFT  & 8x        &  $<1MB$                   &      1                     &  499.2 ms                      \\
DIFT  & 8x        &  $<1MB$                   &      4                     &  2010.4 ms                      \\
RAFT  & 8x         & $56.3 MB$                   &  4                    &   N/A                     \\

\cline{1-5}
\end{tabular}
}
\vspace{-2mm}
\end{table}

 Without further decomposition of cost-volume construction and processing as discussed earlier we cannot run RAFT on-target because of peak-memory constraints i.e., as mobile smartphones have limited (Tightly coupled Memory), often less than 10MB. 
 Instead without decomposition if we try running RAFT on-target based on DDR-memory bound inference we observe that latency of single iteration is less than 0.2 inf/sec. So we do not report latency w.r.t RAFT in our experiments. 
 
We also did not run alternate solutions for optical flow on Snapdragon HTP, because unlike bench-marking on NVIDIA GPUs with CUDA, these mobile platforms ML deployment workflows are still under-development and usually does not support all operations on hardware. Typically such operations have to be identified, substituted and end to end model also might need additional optimizations w.r.t effective usage of hardware for a fair Comparison. As the goal of this study is predominantly to understand to what extent we can adopt cost-volume based solutions for correspondence estimation for mobile, and understand key design choices we start with RAFT and similar intuitions can be extended to various recent follow up works with cost-volumes or cross-attentions, iterative reasoning, etc.

\subsection{Recommendations for Architecture Design:}
\begin{itemize}
    \item As Cost-volume computations are memory-bound, adopt architectures with local cost-volume processing so that overall operation can be decomposed for mobile platforms.
    \item For most settings single-level or coarse-to-fine lookup seem to work with very minimal performance drop with significant boost in latency.
    \item Optimize overall size of cost-volumes by choosing appropriately low resolution of cost-volume at-least for initial iterations to save memory.
    \item Given a choice between number of levels of cost-volumes vs maximum resolution of cost-volume, its better to adopt a maximum resolution one with coarse-to-fine lookup strategy.
    \item Reduce Feature dimension/number of output channels in encoder as much as possible as for many settings within such iterative refinement architectures.
    \item If required increase encoder complexity as it only needs to run-once and doesn't introduce significant bottlenecks w.r.t latency.
    
\end{itemize}



\section{Conclusion} 
In this paper, we have introduced DIFT, a computation and memory efficient real-time optical flow algorithm with competitive performance and good inductive biases . To the best of our knowledge, this is first work to adopt and perform various on-target analysis of such complex cost volume based refinement architectures for mobile use-cases. 
Based on our experiments, we observe that DIFT is feasible for real-time mobile solutions and shows good performance. Nevertheless, any lightweight design choices come at the cost of an expected performance drop. This means, there is a scope for further research to improve correspondence estimation for mobile use-cases as this is a fundamental problem across various vision problems.

\newpage
{\small
\bibliographystyle{ieee_fullname}
\bibliography{egbib}
}

\end{document}


\title{DIFT: Dynamic Iterative Field Transforms for Optical Flow\\ \emph{Supplementary Material}}

\author{First Author\\
Institution1\\
Institution1 address\\
{\tt\small firstauthor@i1.org}
\and
Second Author\\
Institution2\\
First line of institution2 address\\
{\tt\small secondauthor@i2.org}
}

\maketitle

\section{JiT Construction and Lookup}
\begin{figure*}[t]
\begin{center}
\includegraphics[width=\linewidth]{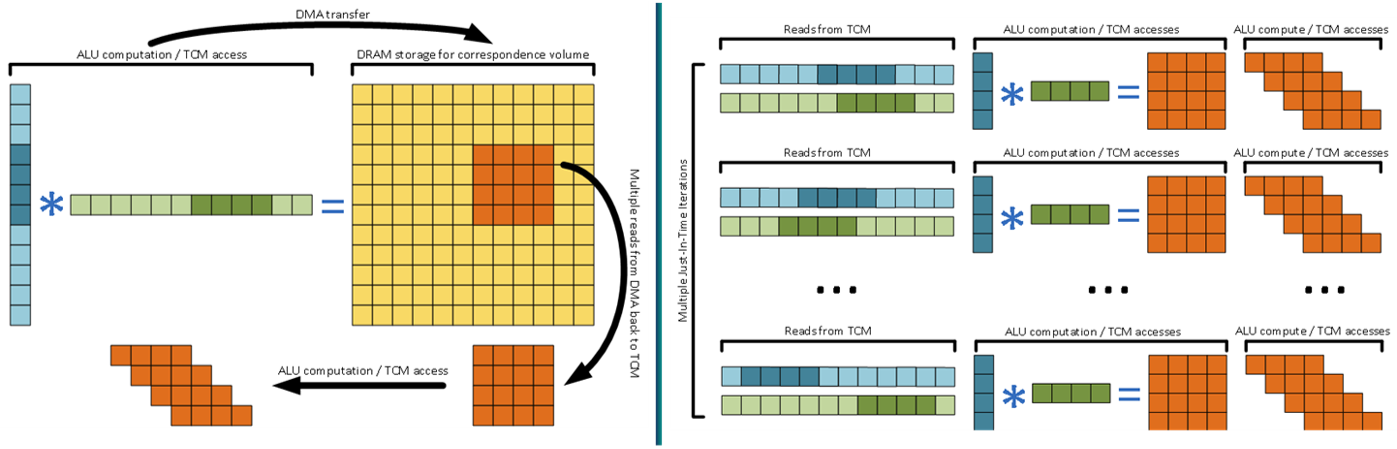}
\end{center}
\caption{JiT Construction  \& Lookup. Blue and green vectors represent feature maps, yellow tensor represnt 4D pre-computed cost volume and orange region represent local neighborhood required for lookup.}
\label{fig:jit}
\end{figure*}

In this section we will review JiT construction and its hardware implications, as illustrated in fig \ref{fig:jit}. As illustrated on the left side, when we pre-compute cost volume visualized by yellow tensor (4D) we will have multiple memory reads in mobile settings as we cannot have entire tensor in-memory (TCM in this case). Hence these multiple reads significantly add to overall latency.

To minimize such memory bottleneck, we compute cost volume only for local neighborhood as illustrated by orange region informed by neighborhood indices based on current estimate of flow. Blue and green vectors illustrate our feature maps(2D) based on which we construct cost volume (4D). So by only querying relevant values we combine cost volume construction and lookup to significanlty reduce peak memory usage and reduce latency.

\section{Coarse to fine Lookup}
\begin{figure*}[t]
\begin{center}
\includegraphics[width=\linewidth]{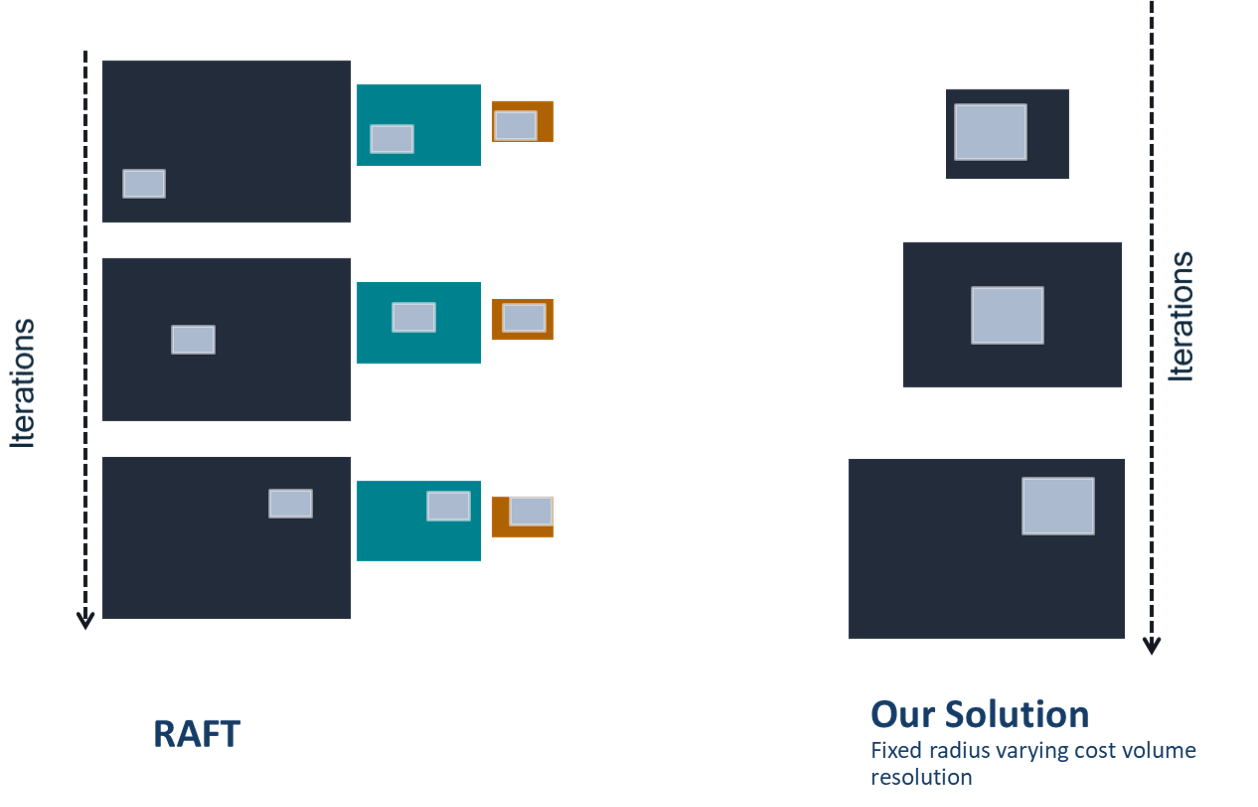}
\end{center}
\caption{Coarse to fine lookup. RAFT's mechanism is illustrated with 3 different resolutions of cost volumes at each iteration, and lookup from \textit{same} location based on current flow estimate for all resolutions within an iteration. Lookup radius is fixed across cost volumes, iterations in case of both RAFT \& Coarse to fine version of DIFT. Resolution of cost volumes is varied in our approach as illustrated on the right}
\label{fig:coarse2fine}
\end{figure*}

In this section we will review coarse to fine lookup with a visual illustration as shown in Fig \ref{fig:coarse2fine}.As illustrated in figure, in case of our reference model RAFT uses multiple levels of cost volumes with varying resolutions to capture coarse to fine information. In case of RAFT they use 4 levels of cost volume but we chose to demonstrate the basic idea with 3 levels of cost volume. Unlike RAFT, our approach within DIFT only uses a single level cost volume with varying resolution of cost volume. 

Given an initial flow estimate (zero in our evaluation setting) RAFT leverages coarse to fine information of cost volumes to have better estimates of flow. But as it is an iterative refinement framework, initial iterations can focus on reaching the neighborhood of true flow and facilitate capturing large displacements. Hence instead of having multiple levels of cost volumes at each iteration we adopt a coarse to fine lookup or cost volume processing where initial iterations would potentially allow network to have a good flow initialization i.e., to be in small neighborhood of true correspondence and then later iterations can focus on finer refinements. 

From Fig  \ref{fig:coarse2fine}, we can observe that for a given pixel RAFT has access to various scales of neighborhood allowing it to capture small \& large motion, allowing more stable refinement. This intuition is validated by our performance boost at lower number of iterations (4 or 6) on both Sintel Clean and Sintel Final settings. We hypothesize performance results on KITTI or more noisy as network was never trained on such large displacement settings and as KITTI only has 200 samples it might not be sufficient data to fine-tune and obtain informative evaluations. It would be very useful to train and evaluate on datasets with large displacements.

\section{Extended Results}

In this section we include extended ablation results for RAFT-small in table \ref{table:Ablation_raft_suppl} and DIFT in table \ref{table:Ablation_total_suppl}.
We include additional ablations varying number of iterations, cost volume levels, feature \& cost volume resolution for RAFT \& DIFT. 

We can observe from \ref{table:Ablation_total_suppl} increasing feature dimension to 256 only marginally improves overall performance compared to 64 but would decrease latency by 4 times.
We would also like to point out that $1/16x$ resolution in case of coarse to fine setting in table \ref{table:Ablation_total_suppl} represents peak resolution or resolution of encoder feature maps within coarse to fine setup, with or without concatenation. We adopt $1/16x, 1/32x, 1/64x$ resolutions in case of coarse to fine setting. 
All other combinations of results also are consistent with our observations in main paper across various settings. 

\begin{table*}[t]
\caption{
Ablation for RAFT on Levels of cost volume, Cost volume resolution, encoder feature dimension and test iterations.
($L^{CV}$: Cost Volume Levels, $R^{CV}$: Cost Volume Resolution (Downsampling), $\#Feat$: Encoder Feature Dimension, and $\#Iter$: Test iterations.)
}
\label{table:Ablation_raft_suppl}
\centering
\adjustbox{max width=\textwidth}
{
\begin{tabular}{|l|cccc|cc|cc|}
\hline
\multirow{2}{*}{Model} & \multirow{2}{*}{$L^{CV}$} & \multirow{2}{*}{$R^{CV}$}  & \multirow{2}{*}{$\#Feat$} & \multirow{2}{*}{$\#Iter$} &  \multicolumn{2}{|c}{Sintel (train-EPE)}  & \multicolumn{2}{|c|}{KITTI (Train)} \\
\cline{6-9}
& & & & & Clean & Final & EPE & F1-all \\

\hline
\multirow{21}{*}{RAFT-small} & \multirow{4}{*}{1}   & \multirow{4}{*}{1/8x}  & \multirow{4}{*}{128} & 4  & 3.434                & 4.624                & 13.522               & 35.804               \\
     &                      &                        &                       & 6  & 2.783                & 3.940                & 10.815               & 31.901               \\
     &                      &                        &                       & 8  & 2.501                & 3.618                & 9.670                & 29.888               \\
     &                      &                        &                       & 12 & 2.340                & 3.422                & 9.062                & 28.201               \\
    \cdashline{2-9}
     & \multirow{4}{*}{2}   & \multirow{4}{*}{1/8x}  &   \multirow{4}{*}{128}                    & 4  & 3.069                & 4.343                & 12.426               & 34.376               \\
     &                      &                        &                       & 6  & 2.606                & 3.813                & 9.953                & 30.390               \\
     &                      &                        &                       & 8  & 2.384                & 3.577                & 8.925                & 28.161               \\
     &                      &                        &                       & 12 & 2.209                & 3.379                & 8.153                & 25.742               \\
     \cdashline{2-9}
     & \multirow{4}{*}{3}   & \multirow{4}{*}{1/8x}  &  \multirow{4}{*}{128}                     & 4  & 2.929                & 4.199                & 12.015               & 34.032               \\
     &                      &                        &                       & 6  & 2.518                & 3.726                & 9.896                & 30.372               \\
     &                      &                        &                       & 8  & 2.332                & 3.529                & 8.892                & 28.172               \\
     &                      &                        &                       & 12 & 2.142                & 3.414                & 8.012                & 25.949               \\
     \cdashline{2-9}
     & \multirow{4}{*}{4}   & \multirow{4}{*}{1/8x}  & \multirow{4}{*}{128}                      & 4  & 2.950                & 4.246                & 11.082               & 33.139               \\
     &                      &                        &                       & 6  & 2.543                & 3.739                & 9.423                & 29.867               \\
     &                      &                        &                       & 8  & 2.360                & 3.496                & 8.708                & 27.995               \\
     &                      &                        &                       & 12 & 2.207                & 3.312                & 8.040                & 26.282               \\
     \cdashline{2-9}
     & \multirow{4}{*}{4}   & \multirow{4}{*}{1/16x} &  \multirow{4}{*}{128}                     & 4  & 3.133                & 4.283                & 12.819               & 43.002               \\
     &                      &                        &                       & 6  & 2.864                & 4.020                & 12.463               & 40.499               \\
     &                      &                        &                       & 8  & 2.754                & 3.939                & 12.240               & 39.093               \\
     &                      &                        &                       & 12 & 2.651                & 3.944                & 11.867               & 38.022               \\
     \cdashline{2-9}
     & \multirow{4}{*}{1}   & \multirow{4}{*}{1/16x} &  \multirow{4}{*}{128}                     & 4  & 3.312                & 4.402                & 12.946               & 45.582               \\
     &                      &                        &                       & 6  & 2.954                & 4.030                & 11.938               & 42.238               \\
     &                      &                        &                       & 8  & 2.797                & 3.851                & 11.319               & 40.126               \\
     &                      &                        &                       & 12 & 2.685                & 3.757                & 10.633               & 38.381              \\

\hline
\end{tabular}
}
\vspace{-3mm}
\centering
\end{table*}


\begin{table*}[t]
\caption{
Ablation for DIFT on Levels of cost volume, Cost volume resolution, encoder feature dimension and test iterations.
($L^{CV}$: Cost Volume Levels, $R^{CV}$: Cost Volume Resolution (Downsampling), $\#Feat$: Encoder Feature Dimension, and $\#Iter$: Test iterations.)
}
\label{table:Ablation_total_suppl}
\centering
\adjustbox{max width=\textwidth}
{
\begin{tabular}{|l|ccc|c|cc|cc|}
\hline
\multirow{2}{*}{Model} & \multirow{2}{*}{$L^{CV}$} & \multirow{2}{*}{$R^{CV}$}  & \multirow{2}{*}{$\#Feat$} & \multirow{2}{*}{$\#Iter$} &  \multicolumn{2}{|c}{Sintel (train-EPE)}  & \multicolumn{2}{|c|}{KITTI (Train)} \\
& & & & & Clean & Final & EPE & F1-all \\

\hline

\cline{2-9}
& \multicolumn{8}{|c|}{Cost Volume Resolution }\\
\cline{2-9}
& \multirow{2}{*}{1} & 1/8 & \multirow{2}{*}{128} & \multirow{2}{*}{12} & 2.70 & 3.93 &10.33  &31.47   \\
&  & 1/16 &  &  & 3.01 & 4.32 & 12.21 & 42.06 \\

\hline

\multirow{27}{*}{DIFT (Ours)}   
     & 1 &  &     &                     & 3.88 & 5.08 & 14.89                   & 47.03                             \\
     & 2 &  &     &                     & 3.84 & 5.10 & 14.51                   & 47.26                             \\
     & 3 & \multirow{-3}{*}{1/16x} & \multirow{-3}{*}{128}     & \multirow{-3}{*}{4} & 3.95 & 5.32 & 13.79 & 44.37          \\
     \cdashline{2-9}
     
     & 1 &  &     &                     & 3.38 & 4.63 & 13.28                   & 44.71                    \\
     & 2 &  &     &                     & 3.42 & 4.54 & 13.79                   & 44.37                            \\
     & 3 & \multirow{-3}{*}{1/16x} & \multirow{-3}{*}{128}     & \multirow{-3}{*}{6} & 3.45 & 4.71 & 13.67                   & 45.88                \\
     \cdashline{2-9}
     & 1 &  &     &                     & 3.01 & 4.32 & 12.21                   & 42.06                           \\
     & 2 &  &     &                     & 3.07 & 4.18 & 12.67                   & 41.15                             \\
 & 3 & \multirow{-3}{*}{1/16x} & \multirow{-3}{*}{128} & \multirow{-3}{*}{12} & 3.10 & 4.17 & 12.54 & 43.07   \\
      \cline{2-9}
      
     & 1 &  &   &                     & 3.85 & 5.03 & 15.12                   & 48.47                              \\
     & 2 &  &   &                     & 3.84 & 5.10 & 14.51                   & 47.26                              \\
     & 3 & \multirow{-3}{*}{1/16x} & \multirow{-3}{*}{64}  & \multirow{-3}{*}{4} & 4.00 & 5.35 & 15.87                   & 48.78      \\
     \cdashline{2-9}
     & 1 &  &   &                     & 3.43 & 4.55 & 13.84                   & 46.07                              \\
     & 2 &  &   &                     & 3.43 & 4.62 & 12.95                   & 44.09                              \\
     & 3 & \multirow{-3}{*}{1/16x} & \multirow{-3}{*}{64}   & \multirow{-3}{*}{6} & 3.51 & 4.79 & 14.48                   & 45.88                              \\
     \cdashline{2-9}
     & 1 &  &   &                     & 3.11 & 4.19 & 12.87                   & 43.83                              \\
     & 2 &  &   &                     & 3.01 & 4.33 & 12.13                   & 42.48                              \\
 & 3 & \multirow{-3}{*}{1/16x} & \multirow{-3}{*}{64}                   & \multirow{-3}{*}{12} & 3.12 & 4.39 & 13.32 & 43.18   \\
  \cline{2-9}
  
     & 1 &  &  &                     & 3.81     & 5.01      & 13.94  & 48.96             \\
     & 2 &  &  &                     & 3.79 & 5.07 & 14.64                   & 49.22                              \\
     & 3 & \multirow{-3}{*}{1/16x} & \multirow{-3}{*}{256}  & \multirow{-3}{*}{4} & 3.74 & 4.88 & 14.63                   & 47.69      \\
     \cdashline{2-9}
     & 1 &  &  &                     & 3.33     & 4.65     & 13.43  & 45.73             \\
     & 2 &  &  &                     & 3.36 & 4.57 & 13.51                   & 46.27                              \\
     & 3 & \multirow{-3}{*}{1/16x} & \multirow{-3}{*}{256}  & \multirow{-3}{*}{6} & 3.32 & 4.49 & 13.39                   & 45.35            \\
     \cdashline{2-9}
     & 1 &  &  &                     & 3.07      & 4.26      & 12.33                        &   43.83                                 \\
     & 2 &  &  &                     & 3.11 & 4.21 & 12.62                   & 42.93                              \\
    & 3 & \multirow{-3}{*}{1/16x} & \multirow{-3}{*}{256}                   & \multirow{-3}{*}{12} & 3.08 & 4.22 & 12.19 & 43.14  \\
\hline
\multirow{4}{*}{Coarse-to-fine (Ours)}   
&  &  &   & 4 & 3.61 & 4.71 & 16.78 & 65.44  \\
&  &  &   & 6 & 3.37 & 4.44 & 14.73 & 56.65  \\
&  &  &   & 8 & 3.33 & 4.42 & 13.56 & 50.82  \\

& \multirow{-4}{*}{varying}   & \multirow{-4}{*}{1/16x}  & \multirow{-4}{*}{128}   
         & 12 & 3.09 & 4.13 & 12.31 & 44.65  \\

\cdashline{1-9}
\multirow{3}{*}{Coarse-to-fine + Concat (Ours)} 
&  &  &   & 4 &3.64  &4.91  &15.12  &54.29   \\
&  &  &   & 6 &3.54  &4.62  &14.85  &51.76   \\

& \multirow{-3}{*}{varying}   & \multirow{-3}{*}{1/16x}  & \multirow{-3}{*}{128}   
        &12 & 3.04 & 4.15 & 13.21 & 47.26 \\

\hline
\end{tabular}
}
\centering
\end{table*}